\pdfoutput=1
\documentclass[11pt]{article}
\usepackage[final]{acl}
\usepackage[T1]{fontenc}
\usepackage[utf8]{inputenc}
\usepackage{times}
\usepackage{latexsym}

\usepackage{amsmath,amsfonts,bm}

\def\eqref#1{equation~\ref{#1}}
\def\1{\bm{1}}

\DeclareMathAlphabet{\mathsfit}{\encodingdefault}{\sfdefault}{m}{sl}
\SetMathAlphabet{\mathsfit}{bold}{\encodingdefault}{\sfdefault}{bx}{n}

\usepackage{graphicx}
\usepackage{tikz}
\usetikzlibrary{arrows.meta,positioning,fit,shapes.geometric}
\usepackage{booktabs}
\usepackage{tabularx}
\usepackage{multirow}
\usepackage{amsmath,amssymb}
\usepackage{subcaption}
\usepackage{algorithm}
\usepackage{algorithmic}
\usepackage{enumitem}
\usepackage{float}
\usepackage{capt-of}
\usepackage[section]{placeins}
\usepackage{microtype}
\usepackage{inconsolata}

\title{StealthRL: Reinforcement Learning Paraphrase Attacks\\for Multi-Detector Evasion of AI-Text Detectors}

\author{
  Suraj Ranganath \\
  University of California, San Diego \\
  \texttt{suranganath@ucsd.edu} \\\And
  Atharv Ramesh \\
  University of California, San Diego \\
  \texttt{a3nair@ucsd.edu}
}

\date{}

\begin{document}

\maketitle

\begin{abstract}
AI-text detectors are increasingly used in high-stakes settings, but their robustness to meaning-preserving adversarial rewriting remains unclear. We introduce \emph{StealthRL}, a reinforcement learning framework for stress-testing detectors with adaptive paraphrase attacks. StealthRL trains a paraphrase policy against a detector ensemble while enforcing semantic fidelity, then evaluates transfer to held-out detector families on the full filtered MAGE test pool. Across RoBERTa, Fast-DetectGPT, Binoculars, and MAGE, the resulting attack sharply degrades detection performance, driving mean TPR@1\%FPR down to 0.024 while retaining nontrivial semantic overlap and readable prose. The attack also transfers to detectors not used during training, indicating a broader robustness failure rather than overfitting to a single target model. We complement the main results with detector-score analysis, repeated-generation stability checks, and automatic and judge-based quality evaluation. Our findings suggest that current AI-text detectors remain brittle under adaptive paraphrasing pressure and motivate robustness evaluation at deployment-relevant operating points. Our code and the trained model checkpoint are publicly available.\footnote{Code: \url{https://github.com/suraj-ranganath/StealthRL}; model: \url{https://huggingface.co/suraj-ranganath/StealthRL}.}
\end{abstract}

% ============================================================================
\section{Introduction}
\label{sec:intro}
% ============================================================================

Large language models (LLMs) produce text that is increasingly indistinguishable from human writing, raising urgent concerns about academic integrity, misinformation, and content provenance~\cite{solaiman2019release}. In response, a growing ecosystem of AI-text detectors has been deployed in educational institutions, publishing platforms, and content moderation systems. These detectors span diverse architectures, from fine-tuned classifiers~\cite{solaiman2019release} to zero-shot statistical methods~\cite{mitchell2023detectgptzeroshotmachinegeneratedtext, bao2024fastdetectgptefficientzeroshotdetection} and paired-LM approaches~\cite{hans2024spottingllmsbinocularszeroshot}, yet their robustness to adversarial manipulation remains poorly understood.

Standard detector evaluations measure performance on \emph{clean} distributions, where AI-generated text is presented without any evasion attempt. However, real-world adversaries are adaptive: they can iteratively refine paraphrases, query detector APIs, and exploit known weaknesses. This gap between clean-distribution evaluation and adversarial robustness is critical. A detector that achieves 95\% accuracy on clean text may fail catastrophically when an attacker deliberately targets its decision boundary.

The choice of operating point further complicates evaluation. Most prior work reports AUROC or accuracy at default thresholds, but deployed detectors in educational and moderation settings are often expected to operate at high precision because false positives can wrongly implicate human writers. This remains true even when detectors are used as human-in-the-loop triage tools rather than autonomous adjudicators: a detector alert still creates review burden, escalation risk, and reputational cost. Commercial systems explicitly discuss false-positive targets below 1\%~\cite{turnitin2023falsepositives}, and recent detector evaluations emphasize TPR measured at fixed FPR values such as 1\%~\cite{tufts-etal-2025-practical}. We therefore treat TPR@1\%FPR as a strict high-precision triage setting, report TPR@5\%FPR as a less conservative review-queue setting, and include AUROC as a threshold-independent view.

We introduce \textbf{StealthRL}, a reinforcement learning framework that systematically evaluates detector robustness under adaptive adversarial conditions. StealthRL trains a paraphrase policy against a multi-detector ensemble, using GRPO~\cite{pang2025grpo} with LoRA adapters~\cite{hu2021loralowrankadaptationlarge} on Qwen3-4B-Instruct, to minimize detection confidence while explicitly rewarding semantic fidelity. By evaluating at the 1\% FPR operating point and testing transfer to multiple held-out detectors, we provide a rigorous robustness assessment that complements standard benchmarks.

Our contributions are as follows:
\begin{itemize}[leftmargin=*,itemsep=2pt]
  \item We implement black-box adaptive paraphrasing attacks via multi-detector RL training with semantic constraints, one of the first works to apply RL for adversarial detector evasion across multiple detector families simultaneously.
  \item We demonstrate catastrophic robustness failure: 0.024 mean TPR@1\%FPR across four detectors, with strong transfer to two held-out detectors.
  \item We provide a reproducible evaluation protocol for measuring evasion, transfer, and quality at security-relevant operating points, along with supporting analyses such as repeat-run stability and semantic evaluation.
\end{itemize}

% ============================================================================
\section{Related Work}
\label{sec:related}
% ============================================================================

\subsection{AI-Text Detection Methods}

AI-text detection methods can be broadly categorized into several families. \textbf{Fine-tuned classifiers} train discriminative models on labeled human and AI text; the RoBERTa-based OpenAI detector~\cite{solaiman2019release} is a widely used example. \textbf{Zero-shot statistical methods} exploit properties of language model probability distributions without requiring labeled data. DetectGPT~\cite{mitchell2023detectgptzeroshotmachinegeneratedtext} uses probability curvature, while Fast-DetectGPT~\cite{bao2024fastdetectgptefficientzeroshotdetection} achieves similar accuracy with substantially reduced computational cost via conditional probability curvature. \textbf{Paired-LM detectors} such as Binoculars~\cite{hans2024spottingllmsbinocularszeroshot} compare log-likelihoods across two language models to detect statistical anomalies. \textbf{Linguistic-feature detectors} use stylometric or language-specific cues rather than only model likelihoods; KatFishNet, for example, detects Korean LLM text using spacing, part-of-speech diversity, and comma-usage features~\cite{park-etal-2025-katfishnet}. The MAGE benchmark~\cite{li2024magemachinegeneratedtextdetection} provides standardized evaluation data, while RAID~\cite{dugan2024raidsharedbenchmarkrobust} offers a shared benchmark for robust detector evaluation across domains and attacks.

Despite diverse architectures, \citet{sadasivan2023aigenerated} provide theoretical arguments that reliable detection may be fundamentally impossible as language models improve, motivating empirical robustness evaluation like ours.

\subsection{Adversarial Attacks on Detectors}

Evasion attacks on AI-text detectors range from simple paraphrasing to sophisticated adaptive methods. \citet{krishna2023paraphrasing} demonstrate that paraphrasing with their DIPPER model effectively evades detectors, though retrieval-based defenses can mitigate the attack. \citet{cheng2025adversarialparaphrasinguniversalattack} study adversarial paraphrasing as a universal attack, using detector-guided candidate selection to humanize AI-generated text. Character-level attacks such as homoglyph substitution (SilverSpeak~\cite{creo2025silverspeakevadingaigeneratedtext}) achieve strong evasion by replacing characters with visually similar Unicode glyphs, but often degrade readability and are detectable by normalization.

Most relevant to our work, AuthorMist~\cite{david2025authormistevadingaitext} applies reinforcement learning to train an evasion policy against a single detector. StealthRL extends this approach to multi-detector ensemble training with held-out evaluation, strict low-FPR operating points, and multi-metric quality evaluation.

Watermark-based detection~\cite{kirchenbauer2023watermark} represents an orthogonal approach where the text generator embeds a statistical signal during generation. While watermarks can provide stronger guarantees, they require control over the generation process and are outside the scope of post-hoc detection methods we evaluate.

A natural defensive response to paraphrasing attacks is adversarial training, where detectors are fine-tuned on adversarially generated examples to improve robustness. However, adversarial training faces a fundamental scalability challenge: the space of possible paraphrasing strategies is vast, and a detector hardened against one attack family may remain vulnerable to novel attack methods. The RAID benchmark~\cite{dugan2024raidsharedbenchmarkrobust} evaluates detectors across diverse attack types and finds that no single detector achieves robust performance against all attacks, underscoring the difficulty of building universally robust defenses. Our work focuses on the attack side of this arms race to quantify the current robustness gap and motivate the development of stronger defensive strategies.

\subsection{Reinforcement Learning for Text Generation}

Reinforcement learning from human feedback (RLHF)~\cite{ouyang2022training} has become the standard approach for aligning language models with human preferences. Proximal Policy Optimization (PPO)~\cite{schulman2017ppo} was the original RL algorithm used for RLHF, but recent work has introduced more efficient alternatives. Group Relative Policy Optimization (GRPO)~\cite{pang2025grpo,shao2024deepseekmath} eliminates the need for a separate value network by using group-level relative rewards, reducing memory requirements and enabling efficient training.

Parameter-efficient fine-tuning methods, particularly LoRA~\cite{hu2021loralowrankadaptationlarge} and QLoRA~\cite{dettmers2023qlora}, enable RL training of large language models with limited compute by adapting only low-rank weight matrices. StealthRL combines GRPO with LoRA on Qwen3-4B-Instruct, demonstrating that efficient RL fine-tuning suffices for learning effective evasion policies.

% ============================================================================
\section{Method}
\label{sec:method}
% ============================================================================

\subsection{Threat Model}
\label{sec:threat}

We consider an adversary who has produced AI-generated text and seeks to modify it to evade detection while preserving semantic content. Formally:

\begin{itemize}[leftmargin=*,itemsep=2pt]
  \item \textbf{Attacker capability}: Black-box access to detector scores. The attacker can query detector confidence $p(y)$ (probability that input $y$ is AI-generated) but does not have access to model gradients or internal parameters. Although our experiments use open-source detectors for which gradients are in principle available, we deliberately treat them as black-box oracles, querying only scalar confidence scores.
  \item \textbf{Attacker goal}: Produce a paraphrase $y$ of AI-generated text $x$ such that $p(y) < \tau$ for detector threshold $\tau$, while maintaining semantic equivalence $\text{sim}(x, y) > \delta$ for a similarity threshold $\delta$.
  \item \textbf{Attacker constraint}: The paraphrase must be a fluent, grammatical reformulation of the original text, not a trivially corrupted version.
\end{itemize}

We evaluate transfer by testing against two \emph{held-out} detectors (Binoculars and MAGE) not seen during training, assessing whether learned evasion strategies generalize across detector architectures.

\begin{figure*}[t]
\centering
\resizebox{\textwidth}{!}{\input{figures/fig_pipeline_vector.tex}}
\caption{StealthRL training and evaluation pipeline. A paraphrase policy (Qwen3-4B with LoRA) is trained via GRPO against a two-detector ensemble (RoBERTa + Fast-DetectGPT) with semantic similarity reward. The trained policy is then evaluated against four detectors, including the held-out Binoculars and MAGE detectors, at the 1\% FPR operating point.}
\label{fig:pipeline}
\end{figure*}

\subsection{Reward Design}
\label{sec:reward}

Given AI-generated text $x$ and paraphrase $y \sim \pi_\theta(\cdot \mid x)$, we define a composite reward:
\begin{equation}
  R(x, y) = \alpha \cdot R_{\text{det}}(y) + \beta \cdot R_{\text{sem}}(x, y),
  \label{eq:reward}
\end{equation}
where $\alpha = 1.0$ and $\beta = 0.1$ control the evasion-quality tradeoff.
We set $\alpha$ to make detector evasion the dominant training signal and use a smaller $\beta$ as a soft fidelity regularizer rather than a competing objective. This choice reflects the attack setting: outputs that remain easily detected fail the primary goal, while semantic preservation is also enforced through the E5 reward, KL regularization, output validation, and downstream quality evaluation.

\paragraph{Detector evasion reward.} The detector reward measures how effectively the paraphrase evades the training ensemble:
\begin{equation}
  R_{\text{det}}(y) = 1 - p_{\text{ens}}(y)
\end{equation}
\begin{equation}
  p_{\text{ens}}(y) = w_1 \cdot p_{\text{RoBERTa}}(y) + w_2 \cdot p_{\text{Fast-DetectGPT}}(y)
\end{equation}
where $w_1 = 0.6$ and $w_2 = 0.4$ are the ensemble weights. The training ensemble intentionally combines one supervised classifier (RoBERTa) and one zero-shot curvature detector (Fast-DetectGPT). We give RoBERTa slightly higher weight because its calibrated neural score provides a stable dense reward, while Fast-DetectGPT contributes a complementary likelihood-curvature signal. We do not include Binoculars or MAGE in the reward loop; they are held out to test transfer, and adding these slower detectors to every GRPO rollout would substantially increase training cost.

Both detector terms are first mapped to an AI-probability scale before averaging. RoBERTa uses the softmax probability assigned to the detector's AI/fake class. Fast-DetectGPT produces an analytic sampling-discrepancy criterion $c(y)$; in the RL reward implementation we transform it as
\begin{equation}
  p_{\text{Fast-DetectGPT}}(y) = \sigma(0.5\,c(y)),
\end{equation}
then clamp the resulting value to $[0,1]$. Thus $p_{\text{ens}}$ is a convex average of two bounded AI-likelihood scores. For GRPO stability, the raw evasion reward $1-p_{\text{ens}}(y)$ is then normalized with a running z-score and clipped to $[-3,3]$ before applying the $\alpha$ coefficient. This normalization is used only inside the RL reward stream; logged detector probabilities and all evaluation metrics use the unnormalized detector scores.

\paragraph{Semantic similarity reward.} To prevent degenerate solutions (e.g., outputting empty or unrelated text), we constrain semantic preservation using E5 embedding cosine similarity~\cite{wang2022e5}:
\begin{equation}
  R_{\text{sem}}(x, y) = \cos\bigl(\text{E5}(x),\; \text{E5}(y)\bigr).
\end{equation}

\paragraph{KL penalty.} GRPO includes an implicit KL divergence penalty (coefficient $\lambda_{\text{KL}} = 0.05$) against the frozen reference policy $\pi_{\text{ref}}$ to prevent catastrophic forgetting and maintain generation fluency.

\subsection{Training Pipeline}
\label{sec:training}

We fine-tune Qwen3-4B-Instruct with LoRA adapters using GRPO~\cite{pang2025grpo}. For each training batch, the policy samples a group of paraphrases, scores each candidate with the detector ensemble and E5 similarity reward, computes group-relative advantages, and updates the LoRA parameters under the KL penalty. The training ensemble comprises RoBERTa and Fast-DetectGPT; Binoculars and the MAGE detector are \emph{held out} for transfer evaluation. Appendix Algorithm~\ref{alg:stealthrl} provides the full pseudocode, and complete training, sampling, and compute settings are reported in Section~\ref{sec:implementation}.

\subsection{Inference}
\label{sec:inference}

At inference time, the fine-tuned policy generates a single paraphrase per input. No candidate selection or reranking is applied; the single-pass output is evaluated directly against all four detectors.

Accordingly, the reported StealthRL attack is \emph{single-shot} at test time: each input receives one policy generation call, with no iterative refinement and no target-detector queries during evaluation. Detector access is required only during offline RL training. This differs from detector-guided search baselines such as M3, which use multi-candidate generation and detector-based reranking at evaluation time.

% ============================================================================
\section{Experimental Setup}
\label{sec:setup}
% ============================================================================

\subsection{Dataset}

We construct a custom subset of the MAGE (Machine-Generated Text Detection in the Wild) benchmark~\cite{li2024magemachinegeneratedtextdetection}, which provides human-written and AI-generated text across multiple domains. Our training set consists of 10,000 AI-generated samples drawn from the MAGE training split; no human-written samples are used during RL fine-tuning, as the policy learns entirely from detector feedback on paraphrased AI text. The evaluation set uses the full filtered MAGE test pool with 15,310 human-written and 14,656 AI-generated samples after applying the 100--500 token filter. Human samples in the evaluation set are used solely for threshold calibration at the 1\% FPR operating point; AI and attacked samples are never used when setting thresholds.

\subsection{Detectors}

We evaluate against four detectors spanning distinct architectural paradigms:

\begin{itemize}[leftmargin=*,itemsep=2pt]
  \item \textbf{RoBERTa OpenAI} (\texttt{openai-community/roberta-large-}\\\texttt{openai-detector}): A fine-tuned RoBERTa-large classifier trained on GPT-2 outputs. Represents the class of supervised neural classifiers.
  \item \textbf{Fast-DetectGPT}~\cite{bao2024fastdetectgptefficientzeroshotdetection}: A zero-shot statistical method that uses conditional probability curvature with scoring model \texttt{EleutherAI/gpt-neo-2.7B}. Represents curvature-based detection.
  \item \textbf{Binoculars}~\cite{hans2024spottingllmsbinocularszeroshot}: A paired-LM detector using \texttt{gpt2-medium} and \texttt{gpt2-large} that compares cross-entropy and perplexity. \emph{Held out during training} to assess transfer.
  \item \textbf{MAGE detector}~\cite{li2024magemachinegeneratedtextdetection}: The benchmark-provided Longformer-based classifier released with MAGE (model checkpoint \texttt{yaful/MAGE}). In the original MAGE paper, this PLM-based detector is the strongest method among their compared detectors across the benchmark's in-distribution and out-of-distribution testbeds, so we include it as a strong held-out evaluation target rather than a detector used during training.
\end{itemize}

\subsection{Baselines}

We compare six attack settings:
\begin{itemize}[leftmargin=*,itemsep=2pt]
  \item \textbf{M0 (No Attack)}: Original AI-generated text, unmodified.
  \item \textbf{M1 (Simple Paraphrase)}: Standard LLM paraphrasing without detector feedback.
  \item \textbf{M2 (StealthRL)}: Our method, RL-trained paraphrasing against a detector ensemble (3 epochs, GRPO + LoRA).
  \item \textbf{M3 (Adversarial Paraphrasing)}~\cite{cheng2025adversarialparaphrasinguniversalattack}: Detector-guided candidate selection from multiple paraphrases.
  \item \textbf{M4 (AuthorMist)}~\cite{david2025authormistevadingaitext}: RL-trained paraphrasing against a single detector (RoBERTa).
  \item \textbf{M5 (Homoglyph)}~\cite{creo2025silverspeakevadingaigeneratedtext}: Character-level Unicode substitution (SilverSpeak).
\end{itemize}

Only M3 uses detector-guided search at evaluation time: it samples four candidates and reranks them using RoBERTa subject to an E5 similarity floor. StealthRL (M2) is evaluated as a single-shot policy with one generation per input and no detector-based reranking or evaluation-time detector queries. Appendix Tables~\ref{tab:attack_protocol} and~\ref{tab:attack_protocol_queries} report the complete per-method query budget, candidate count, generator, prompt/decoding settings, reranking detector, and seed policy.

\subsection{Implementation Details}
\label{sec:implementation}

StealthRL fine-tunes Qwen3-4B-Instruct with LoRA adapters (rank 32, LoRA $\alpha=32$, dropout 0.05) for 3 epochs on 10,000 AI-generated MAGE training samples. GRPO uses batch size 16, group size 8, learning rate $2.8 \times 10^{-4}$, maximum 512 generated tokens, and KL coefficient $\lambda_{\text{KL}}=0.05$ against the frozen reference policy. The inference prompt is \texttt{``Paraphrase the following text while preserving its meaning: [TEXT]''}; decoding uses temperature 1.0 and top-p 0.9. To test whether results depend on one stochastic draw, we additionally run two full-pool M2 repeats with the same checkpoint, prompt, and decoding configuration but different random seeds (4201 and 4202), scoring each repeat with the same detector pipeline and fixed thresholds.

The training reward uses RoBERTa and Fast-DetectGPT because they expose complementary detector signals while remaining fast enough for repeated reward calls inside GRPO. Binoculars and MAGE are excluded from training and reserved for held-out transfer evaluation. This design makes the evaluation stricter, but it also means our training ensemble is not exhaustive; scaling RL to larger and slower detector panels remains future work.

Appendix~\ref{app:ablations} reports the small-subset ablations used for model selection. These ablations were run on a 1,000-human / 1,000-AI MAGE selection subset and were used to choose the final reward setting before full-test evaluation; all main results in Section~\ref{sec:results} are recomputed on the full filtered MAGE test pool.

\subsection{Evaluation Metrics}

We report detection metrics and quality diagnostics:
\begin{itemize}[leftmargin=*,itemsep=2pt]
  \item \textbf{TPR@1\%FPR}: True positive rate at 1\% false positive rate, using a detector threshold calibrated as the empirical 99th percentile of scores on the 15,310 human evaluation samples. This is our primary high-precision operating point.
  \item \textbf{TPR@5\%FPR}: The same metric at a looser 5\% false positive rate, using the empirical 95th percentile of human scores. We report this as a supplementary operating point.
  \item \textbf{ASR} (Attack Success Rate): Fraction of AI samples classified as human at the 1\% FPR threshold. $\text{ASR} = 1 - \text{TPR@1\%FPR}$.
  \item \textbf{AUROC}: Area under the receiver operating characteristic curve. Threshold-independent measure of overall discriminability.
  \item \textbf{E5 Similarity}: Cosine similarity between E5 embeddings~\cite{wang2022e5} of original and paraphrased text, measuring sentence-level semantic preservation.
  \item \textbf{BERTScore}: Contextual token-alignment F1~\cite{zhang2020bertscore}, which complements E5 by measuring local lexical and semantic overlap rather than only global embedding similarity.
  \item \textbf{NLI Consistency}: Bidirectional entailment consistency using a DeBERTa-large model fine-tuned on MNLI~\cite{he2021deberta}. We score both original$\rightarrow$paraphrase and paraphrase$\rightarrow$original directions and report the minimum entailment probability, so a high value requires meaning preservation in both directions.
  \item \textbf{Perplexity (PPL)}: GPT-2 perplexity of the output text. Lower values indicate lower language-model surprise under GPT-2; PPL is reported as a fluency diagnostic, not as a higher-is-better quality score.
  \item \textbf{Edit Rate}: $1-$\texttt{SequenceMatcher} character-level similarity between the source and output. It measures how much text changed; higher values are not inherently better because both trivial copying and excessive rewriting can be undesirable.
  \item \textbf{Valid}: Fraction of outputs passing the automatic validity filter: non-empty, at least 10 words, and no more than $3\times$ the original character length.
\end{itemize}

All confidence intervals are computed via bootstrap resampling with 500 iterations and seed 42.

\subsection{LLM-Based Quality Judge}

We additionally evaluate paraphrase quality with an LLM judge, following prior work on automatic text-quality evaluation~\cite{fu2023gptscore,kim2024prometheus2,vu2024foundationalautoraters}. OpenAI \texttt{gpt-5-nano} scores each paraphrase on two 1--5 Likert axes: linguistic quality and semantic similarity. For fair cross-method comparison, we score a shared subset of 500 AI samples per method (M1--M5) using identical sample IDs; the judge sees only the source and paraphrase, without method labels. Appendix~\ref{app:prompts} reports the exact prompts.

% ============================================================================
\section{Results}
\label{sec:results}
% ============================================================================

\subsection{Main Detection Evasion Results}

StealthRL (M2) achieves a \textbf{0.024 mean TPR@1\%FPR}, reducing mean AUROC from 0.79 (no attack) to 0.43 with a 97.6\% attack success rate. At this strict high-precision triage setting, the evaluated detector panel becomes largely ineffective against the adaptive attack. The full detector-by-detector AUC/TPR/ASR tables for the 1\% FPR operating point are moved to Appendix Tables~\ref{tab:app_main_binoc}, \ref{tab:app_main_fastd}, \ref{tab:app_main_mage}, and~\ref{tab:app_main_part2}.

\begin{figure*}[t]
  \centering
  \includegraphics[width=0.98\textwidth]{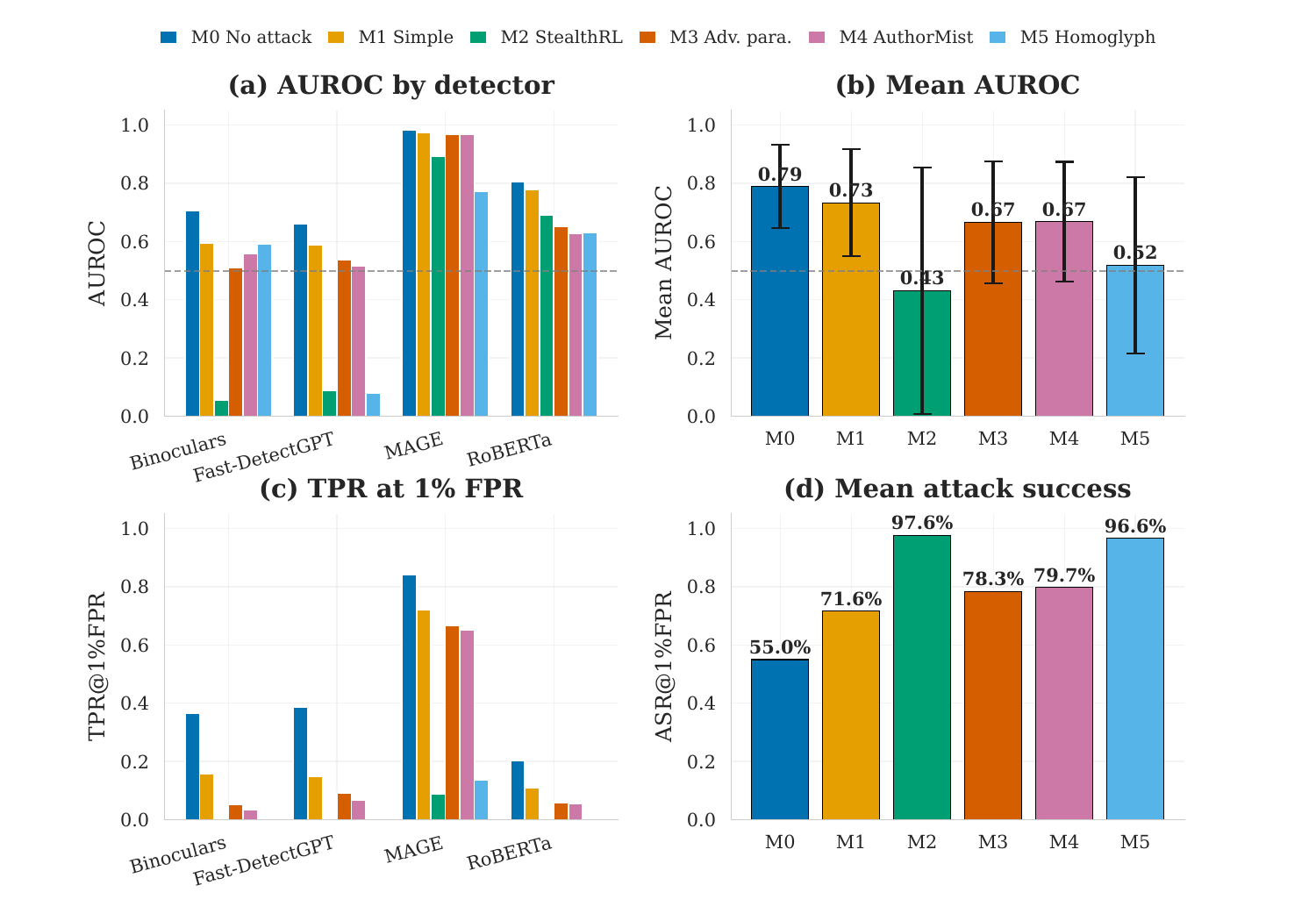}
  \caption{Detection evasion results for methods M0--M5. (a) AUROC by detector. (b)~Mean AUROC with standard-deviation bars across detectors. (c) TPR at 1\% FPR. (d) Mean attack success rate. StealthRL (M2, teal) achieves extremely low TPR on Binoculars, Fast-DetectGPT, and RoBERTa, with MAGE remaining the most robust detector in the panel.}
  \label{fig:method_comparison}
\end{figure*}

Figure~\ref{fig:method_comparison} summarizes the main comparison across metrics. Panel~(a) shows that StealthRL dramatically reduces AUROC across the detector panel, driving Binoculars and Fast-DetectGPT close to random chance and substantially degrading the remaining detectors. Panel~(b) confirms that the mean AUROC of 0.432 for M2 is far below the no-attack baseline of 0.789. Panel~(c) demonstrates near-zero TPR at strict operating points on three detectors, while Panel~(d) shows the 97.6\% attack success rate.

\begin{figure}[t]
  \centering
  \includegraphics[width=0.86\columnwidth]{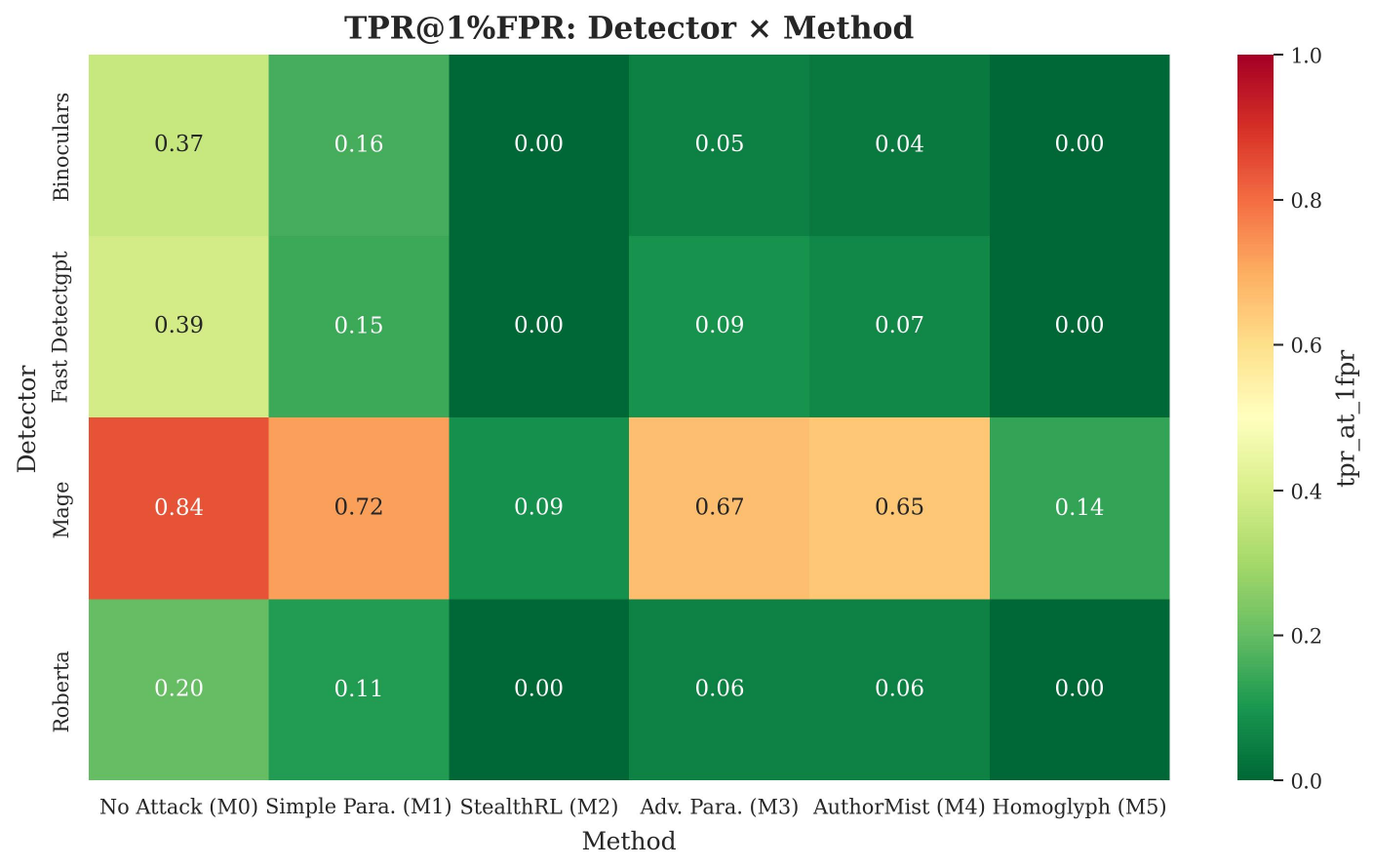}
  \caption{TPR@1\%FPR heatmap. Lower is better for the attacker; M2 collapses TPR on three detectors and degrades held-out MAGE.}
  \label{fig:heatmap}
\end{figure}

\subsection{Robustness, Transfer, and Quality}

StealthRL's evasion transfers to detectors not used during training: M2 reaches 0.002 TPR@1\%FPR on held-out Binoculars and 0.089 on the held-out MAGE detector, alongside 0.002 on both in-ensemble detectors. The same ranking holds at the supplementary 5\% FPR operating point, where M2 obtains 0.076 mean TPR@5\%FPR (Appendix Table~\ref{tab:app_tpr5}). Repeated stochastic generation is also stable: across the original run and two full-pool repeats, mean AUROC is 0.432 $\pm$ 0.001, mean TPR@1\%FPR is 0.024 $\pm$ 0.001, and mean ASR is 0.976 $\pm$ 0.001 (Appendix Table~\ref{tab:stochastic_repeats}).

Figure~\ref{fig:heatmap} and Appendix Figure~\ref{fig:app_score_dist} provide the per-detector heatmap and score distributions. The distributions show that M2 pushes most AI-sample scores below the calibrated 1\% FPR threshold, explaining why TPR collapses even when a detector such as RoBERTa retains some global rank information. This pattern is consistent with cross-detector vulnerabilities within our panel, but we do not claim a universal mechanism for all detector families.

The attack also exposes an evasion--fidelity tradeoff. M2 obtains near-maximal mean ASR (0.976) with E5 similarity 0.901, BERTScore 0.868, NLI 0.316, and judged quality 2.51 (Appendix Table~\ref{tab:quality}). Thus, M2 is more readable than the homoglyph baseline (M5 quality 1.97) but less semantically faithful than high-quality paraphrasing baselines M1/M3. We therefore interpret StealthRL as a robustness stress test rather than a quality-equivalent paraphraser.

\subsection{Discussion}

These results show that, for the evaluated detector panel, adaptive paraphrasing can defeat high-precision detector-assisted review workflows with one generation per input and no evaluation-time detector reranking. Human review changes how detector alerts should be used, but it does not remove the need for robust triage signals: low precision overloads reviewers, while low TPR fails to surface adversarially paraphrased AI text for review. Purpose-built adversarial training, provenance-aware defenses, and more semantically grounded detectors are therefore necessary next steps.

% ============================================================================
\section{Conclusion and Future Work}
\label{sec:conclusion}
% ============================================================================

StealthRL demonstrates severe detector failure under adaptive RL-based paraphrasing attacks, revealing significant robustness gaps in the AI-text detectors evaluated. By training against a multi-detector ensemble with GRPO and LoRA, we achieve near-zero detection on three of four detectors and a 0.024 mean TPR@1\%FPR with strong cross-architecture transfer, including to two held-out detectors. Our evaluation, spanning detection metrics, stochastic repeat analysis, multi-metric quality assessment, and score distribution analysis, characterizes the evasion--quality tradeoff on this benchmark.

Future work should study adversarial training, provenance-aware defenses, semantic and discourse-level detection signals, stronger constrained RL objectives, and broader datasets, languages, and detector families. Our evaluation framework and released code provide a rigorous testbed for measuring progress on adversarially robust AI-text detection.

% ============================================================================
\section{Limitations}
\label{sec:limitations}
% ============================================================================

\paragraph{Detector coverage.} Our evaluation covers four detectors spanning supervised, zero-shot statistical, paired-LM, and benchmark-provided paradigms. We do not evaluate against watermark-based detectors~\cite{kirchenbauer2023watermark}, which embed signals during generation and may be more robust to paraphrasing attacks. Evaluating StealthRL against watermarked text is an important direction for future work.

\paragraph{Training-ensemble coverage.} The RL reward uses two detectors, RoBERTa and Fast-DetectGPT, while Binoculars and MAGE are held out for transfer evaluation. This separation is useful for measuring cross-detector transfer, but it also means we do not study whether training directly against a larger detector ensemble would further improve or destabilize evasion. Larger ensembles would require substantially more reward-model calls per GRPO update, especially for paired-LM and Longformer-based detectors, and are left for future work.

\paragraph{Dataset diversity.} We evaluate on a single benchmark (MAGE) in English. Broader coverage across datasets (RAID~\cite{dugan2024raidsharedbenchmarkrobust}), domains, and languages is needed to establish generalizability. The MAGE benchmark, while diverse, may not capture all deployment scenarios.

\paragraph{Quality gap.} StealthRL achieves lower semantic fidelity than the high-quality paraphrase baselines (E5 0.901, BERTScore 0.868, and NLI 0.316, compared with M1/M3 E5 0.974/0.973, BERTScore 0.926/0.934, and NLI 0.967/0.965) and lower Likert quality (2.51 vs.\ 3.78/3.77). It still outperforms character-level obfuscation on BERTScore and judged quality (M5: 0.773 and 1.97), but the remaining fidelity cost is a limitation of the current system. We supplement E5 with BERTScore, bidirectional NLI, and LLM-based similarity judgments, but all automatic semantic metrics remain imperfect proxies for human meaning preservation. Improving semantic preservation while maintaining strong evasion is an important direction. Techniques such as constrained decoding, rejection sampling, or multi-objective RL could help close this gap.

\paragraph{Defense evaluation.} We do not explore defensive strategies such as adversarial training, certified robustness, or ensemble diversification that could improve detector resilience. Our focus is on exposing vulnerabilities to motivate defensive research.

% ============================================================================
\section{Ethical Considerations}
% ============================================================================

Adversarial paraphrasing is dual-use technology. We position StealthRL as a \emph{stress-testing and robustness evaluation tool} for researchers and detector developers, not a production evasion system. The near-zero TPR@1\%FPR result exposes critical vulnerabilities that must be addressed before detectors are deployed in high-stakes applications such as academic integrity enforcement.

We release our code and evaluation pipeline to enable reproducible robustness assessment and to accelerate defensive research. By making attack capabilities transparent, we aim to shift the detector development paradigm toward adversarial robustness rather than clean-distribution accuracy. We believe responsible disclosure of detector vulnerabilities, accompanied by tools for measuring progress, serves the broader goal of trustworthy AI-text detection.

% ============================================================================
\section*{Acknowledgments}
% ============================================================================

We gratefully acknowledge Thinking Machines for providing free research credits and access to their Tinker API framework, which made the RL fine-tuning possible. We also thank the open-source community for the detector implementations and model checkpoints that enabled this evaluation.

\bibliography{stealthrl_refs}
\clearpage

% ============================================================================
\appendix
% ============================================================================

\section{Training Algorithm}
\label{app:algorithm}

\begin{algorithm}[H]
\caption{StealthRL Training}
\label{alg:stealthrl}
\begin{algorithmic}[1]
\REQUIRE Training data $\mathcal{D} = \{x_i\}_{i=1}^N$ (AI-generated texts), detector ensemble $\{d_k\}_{k=1}^K$ with weights $\{w_k\}$, E5 similarity model, base LLM $\pi_{\text{ref}}$
\ENSURE Fine-tuned paraphrase policy $\pi_\theta$
\STATE Initialize $\pi_\theta \leftarrow \pi_{\text{ref}}$ with LoRA adapters (rank $r{=}32$, $\alpha{=}32$)
\FOR{epoch $= 1, \ldots, E$}
  \FOR{batch $\mathcal{B} \subset \mathcal{D}$}
    \FOR{each $x \in \mathcal{B}$}
      \STATE Sample group $\{y_1, \ldots, y_G\} \sim \pi_\theta(\cdot \mid x)$ \hfill $\triangleright$ $G{=}8$
      \FOR{each $y_g$ in group}
        \STATE $R_{\text{det}}(y_g) \leftarrow 1 - \sum_{k} w_k \cdot d_k(y_g)$
        \STATE $R_{\text{sem}}(x, y_g) \leftarrow \cos(\text{E5}(x), \text{E5}(y_g))$
        \STATE $R(x, y_g) \leftarrow \alpha R_{\text{det}}(y_g) + \beta R_{\text{sem}}(x, y_g)$
      \ENDFOR
      \STATE Compute group-relative advantages: $A_g \leftarrow (R_g - \bar{R}) / \sigma_R$
    \ENDFOR
    \STATE Update $\theta$ via clipped policy gradient with KL penalty
  \ENDFOR
\ENDFOR
\RETURN $\pi_\theta$
\end{algorithmic}
\end{algorithm}

\FloatBarrier
\section{Full Main Results on the MAGE Test Pool}
\label{app:mainresults}

\begin{center}
\small
\setlength{\tabcolsep}{4pt}
\resizebox{\columnwidth}{!}{%
  \begin{tabular}{lrrr}
\toprule
Method & Binoc.\ AUC & Binoc.\ TPR & Binoc.\ ASR \\
\midrule
M0 & 0.705 & 0.367 & 0.633 \\
M1 & 0.593 & 0.158 & 0.842 \\
M2 (Ours) & \textbf{0.055} & 0.002 & 0.998 \\
M3 & 0.511 & 0.051 & 0.949 \\
M4 & 0.559 & 0.036 & 0.964 \\
M5 & 0.593 & \textbf{0.000} & \textbf{1.000} \\
\bottomrule
\end{tabular}

}
\captionof{table}{Main results on the full filtered MAGE test pool at the 1\% FPR operating point for Binoculars. Lower TPR@1\%FPR/AUROC is better for the attacker; higher ASR is better. Bold indicates the best value per metric.}
\label{tab:app_main_binoc}
\end{center}

\begin{center}
\small
\setlength{\tabcolsep}{4pt}
\resizebox{\columnwidth}{!}{%
  \begin{tabular}{lrrr}
\toprule
Method & FastDGPT AUC & FastDGPT TPR & FastDGPT ASR \\
\midrule
M0 & 0.661 & 0.388 & 0.612 \\
M1 & 0.588 & 0.148 & 0.852 \\
M2 (Ours) & 0.089 & 0.002 & 0.998 \\
M3 & 0.536 & 0.092 & 0.908 \\
M4 & 0.518 & 0.067 & 0.933 \\
M5 & \textbf{0.080} & \textbf{0.000} & \textbf{1.000} \\
\bottomrule
\end{tabular}

}
\captionof{table}{Main results on the full filtered MAGE test pool at the 1\% FPR operating point for Fast-DetectGPT. Lower TPR@1\%FPR/AUROC is better for the attacker; higher ASR is better. Bold indicates the best value per metric.}
\label{tab:app_main_fastd}
\end{center}

\begin{center}
\small
\setlength{\tabcolsep}{4pt}
\resizebox{\columnwidth}{!}{%
  \begin{tabular}{lrrr}
\toprule
Method & MAGE AUC & MAGE TPR & MAGE ASR \\
\midrule
M0 & 0.983 & 0.843 & 0.157 \\
M1 & 0.973 & 0.720 & 0.280 \\
M2 (Ours) & 0.891 & \textbf{0.089} & \textbf{0.911} \\
M3 & 0.967 & 0.666 & 0.334 \\
M4 & 0.969 & 0.651 & 0.349 \\
M5 & \textbf{0.772} & 0.136 & 0.864 \\
\bottomrule
\end{tabular}

}
\captionof{table}{Main results on the full filtered MAGE test pool at the 1\% FPR operating point for MAGE. Lower TPR@1\%FPR/AUROC is better for the attacker; higher ASR is better. Bold indicates the best value per metric.}
\label{tab:app_main_mage}
\end{center}

\begin{center}
\small
\setlength{\tabcolsep}{4pt}
\resizebox{\columnwidth}{!}{%
  \begin{tabular}{lrrrrr}
\toprule
Method & RoBERTa AUC & RoBERTa TPR & RoBERTa ASR & Mean TPR & Mean ASR \\
\midrule
M0 & 0.806 & 0.203 & 0.797 & 0.450 & 0.550 \\
M1 & 0.778 & 0.111 & 0.889 & 0.284 & 0.716 \\
M2 (Ours) & 0.691 & 0.002 & 0.998 & \textbf{0.024} & \textbf{0.976} \\
M3 & 0.651 & 0.058 & 0.942 & 0.217 & 0.783 \\
M4 & \textbf{0.627} & 0.056 & 0.944 & 0.203 & 0.797 \\
M5 & 0.630 & \textbf{0.001} & \textbf{0.999} & 0.034 & 0.966 \\
\bottomrule
\end{tabular}

}
\captionof{table}{Main results on the full filtered MAGE test pool at the 1\% FPR operating point, part II. Lower TPR@1\%FPR/AUROC is better for the attacker; higher ASR is better. This panel reports RoBERTa and the mean TPR@1\%FPR / mean ASR aggregates across the four-detector panel. Bold indicates the best value per metric.}
\label{tab:app_main_part2}
\end{center}

\begin{center}
\small
\setlength{\tabcolsep}{5pt}
\resizebox{\columnwidth}{!}{%
  \begin{tabular}{lrrrrr}
\toprule
Method & Binoc.\ TPR@5\% & FastDGPT TPR@5\% & MAGE TPR@5\% & RoBERTa TPR@5\% & Mean TPR@5\% \\
\midrule
M0 & 0.481 & 0.451 & 0.934 & 0.397 & 0.566 \\
M1 & 0.269 & 0.252 & 0.876 & 0.294 & 0.423 \\
M2 (Ours) & 0.005 & 0.003 & 0.272 & \textbf{0.022} & 0.076 \\
M3 & 0.135 & 0.180 & 0.846 & 0.159 & 0.330 \\
M4 & 0.126 & 0.130 & 0.839 & 0.158 & 0.313 \\
M5 & \textbf{0.004} & \textbf{0.000} & \textbf{0.238} & 0.027 & \textbf{0.067} \\
\bottomrule
\end{tabular}

}
\captionof{table}{Supplementary TPR@5\%FPR results on the full filtered MAGE test pool. Lower is better for the attacker. Thresholds are recalibrated on the 15,310 human evaluation samples at the 5\% false-positive operating point. Bold indicates the lowest TPR@5\%FPR per column.}
\label{tab:app_tpr5}
\end{center}

\begin{center}
\small
\setlength{\tabcolsep}{3pt}
\resizebox{\columnwidth}{!}{%
  \begin{tabular}{lrrrrrrr}
\toprule
Run & Mean AUROC & Mean TPR & Mean ASR & Binoc. & Fast-DGPT & MAGE & RoBERTa \\
\midrule
Original M2 & 0.432 & 0.024 & 0.976 & 0.002 & 0.002 & 0.089 & 0.002 \\
Seed 4201 & 0.432 & 0.025 & 0.975 & 0.002 & 0.002 & 0.093 & 0.002 \\
Seed 4202 & 0.433 & 0.024 & 0.976 & 0.002 & 0.002 & 0.091 & 0.002 \\
\midrule
Mean $\pm$ SD & 0.432 $\pm$ 0.001 & 0.024 $\pm$ 0.001 & 0.976 $\pm$ 0.001 & 0.002 $\pm$ 0.000 & 0.002 $\pm$ 0.000 & 0.091 $\pm$ 0.002 & 0.002 $\pm$ 0.000 \\
\bottomrule
\end{tabular}

}
\captionof{table}{Full-pool stochastic repeat analysis for StealthRL (M2) at temperature 1.0 and top-p 0.9. All runs use one generation per input and no reranking; thresholds are the same 1\% FPR thresholds calibrated on human evaluation samples. Detector columns report TPR@1\%FPR.}
\label{tab:stochastic_repeats}
\end{center}

\FloatBarrier
\section{Quality Metrics}
\label{app:quality}

\begin{center}
  \small
  \setlength{\tabcolsep}{6pt}
  \resizebox{\columnwidth}{!}{\begin{tabular}{lrrrrrrrr}
\toprule
Method & E5 Sim. & BERTScore & NLI & PPL & Edit Rate & Valid & Quality & Similarity \\
\midrule
M0 & 1.000 & 1.000 & 0.991 & 26.662 & 0.000 & 1.000 & --- & --- \\
M1 & 0.974 & 0.926 & 0.967 & 24.596 & 0.779 & 1.000 & 3.780 & 4.058 \\
M2 (Ours) & 0.901 & 0.868 & 0.316 & 148.686 & 0.835 & 1.000 & 2.514 & 2.644 \\
M3 & 0.973 & 0.934 & 0.965 & 28.338 & 0.782 & 1.000 & 3.770 & 4.024 \\
M4 & 0.958 & 0.902 & 0.829 & 18.909 & 0.861 & 0.998 & 3.152 & 3.704 \\
M5 & 0.923 & 0.773 & 0.847 & 146.430 & 0.634 & 1.000 & 1.972 & 2.920 \\
\bottomrule
\end{tabular}
}
  \captionof{table}{Quality and similarity metrics. E5 Sim., BERTScore, and NLI are computed on the full filtered evaluation pool. Quality and Similarity are mean Likert scores (1--5) from a gpt-5-nano judge on 500 matched samples for methods M1--M5; M0 reports automatic metrics only. Higher is better for E5, BERTScore, NLI, Valid, Quality, and Similarity. Lower PPL indicates lower GPT-2 language-model surprise, while Edit Rate is a magnitude-of-change diagnostic rather than a quality-positive score.}
  \label{tab:quality}
\end{center}

\FloatBarrier
\section{Qualitative Examples}
\label{app:examples}

Tables~\ref{tab:qual1}--\ref{tab:qual3} show representative paraphrases across different domains from the filtered MAGE test split.

\begin{center}
\footnotesize
\setlength{\tabcolsep}{4pt}
\begin{tabularx}{\columnwidth}{>{\raggedright\arraybackslash}X >{\raggedright\arraybackslash}X}
\toprule
\textbf{Original (M0)} & \textbf{StealthRL (M2)} \\
\midrule
During cardio the heart increases its workload and all the body's other systems adjust to help support that endeavor. The blood vessels dilate, the muscles do their best to help pump blood back to the heart, and the lungs work harder to take in oxygen and remove waste gases like carbon dioxide. &
In cardio, the heart ramps up its workload, prompting the body's other systems to adapt and assist. Blood vessels widen, muscles strive to push blood back to the heart, and lungs intensify their efforts to absorb oxygen and expel waste gases such as carbon dioxide. \\
\midrule
The engine has to endure the torque of powering two axels and a drive shaft generally the transfer casing connects to the drive shaft with a ujoint and same for power distribution. The electricity goes into the battery first then is sent to the alternator where it generates voltage that powers all other electrical components like the lights, radio. &
The engine must handle the torque from two axles and a drive shaft, typically linked via a universal joint in the transfer casing for power distribution. Electricity first charges the battery, then flows to the alternator, which converts it into voltage to power essential electrical systems such as lights and the radio. \\
\bottomrule
\end{tabularx}
\captionof{table}{Representative paraphrases from the filtered MAGE test split (health and automotive domains).}
\label{tab:qual1}
\end{center}

\begin{center}
\footnotesize
\setlength{\tabcolsep}{4pt}
\begin{tabularx}{\columnwidth}{>{\raggedright\arraybackslash}X >{\raggedright\arraybackslash}X}
\toprule
\textbf{Original (M0)} & \textbf{StealthRL (M2)} \\
\midrule
Machine learning algorithms have become increasingly sophisticated in recent years, with deep neural networks achieving state-of-the-art performance across a wide range of tasks including image classification, natural language processing, and reinforcement learning. &
Recent advances in machine learning have led to highly sophisticated algorithms. Deep neural networks now achieve top performance in diverse areas, from classifying images and processing natural language to learning through reinforcement-based approaches. \\
\midrule
The process of photosynthesis converts light energy into chemical energy that can be stored in glucose molecules. This process occurs in the chloroplasts of plant cells and involves two main stages: the light-dependent reactions and the Calvin cycle. &
Photosynthesis transforms light energy into chemical energy, storing it within glucose molecules. Taking place in plant cell chloroplasts, this process unfolds in two key stages: light-dependent reactions followed by the Calvin cycle. \\
\bottomrule
\end{tabularx}
\captionof{table}{Representative paraphrases (science and technology domains).}
\label{tab:qual2}
\end{center}

\begin{center}
\footnotesize
\setlength{\tabcolsep}{4pt}
\begin{tabularx}{\columnwidth}{>{\raggedright\arraybackslash}X >{\raggedright\arraybackslash}X}
\toprule
\textbf{Original (M0)} & \textbf{StealthRL (M2)} \\
\midrule
The global economy faces several interconnected challenges, including rising inflation, supply chain disruptions, and geopolitical tensions. Central banks worldwide have responded by tightening monetary policy, raising interest rates to combat price pressures. &
Several intertwined challenges confront the global economy: climbing inflation, supply chain breakdowns, and geopolitical friction. In response, central banks around the world have tightened monetary policy and hiked interest rates to curb rising prices. \\
\midrule
Renewable energy sources such as solar and wind power have seen dramatic cost reductions over the past decade, making them competitive with traditional fossil fuels in many markets. Government incentives and technological improvements continue to drive adoption rates higher. &
Solar and wind power costs have fallen dramatically over the last decade, allowing renewables to compete with fossil fuels across numerous markets. Ongoing government incentives and technological progress continue pushing adoption rates upward. \\
\bottomrule
\end{tabularx}
\captionof{table}{Representative paraphrases (economics and energy domains).}
\label{tab:qual3}
\end{center}

\FloatBarrier
\section{Attack Protocol and Query Budgets}
\label{app:protocol}

\noindent\textbf{Prompt templates.} M1 and M3 use the standard prompt \texttt{``Please paraphrase the following text while maintaining its meaning and style. Keep the paraphrase close to the original length, do not add new details, and output only the paraphrased text without any additional explanation.''} M2 uses the StealthRL prompt reported in Table~\ref{tab:hyperparams}. M4 uses the AuthorMist humanization prompt \texttt{``Please paraphrase the following text to make it more human-like while preserving the original meaning. Keep the paraphrase close to the original length and do not add new details.''} M0 and M5 use no LLM prompt.

\begin{center}
\scriptsize
\setlength{\tabcolsep}{2pt}
\begin{tabularx}{\columnwidth}{l c >{\raggedright\arraybackslash}X >{\raggedright\arraybackslash}X}
\toprule
\textbf{Method} & \textbf{Cand.} & \textbf{Generator / transform} & \textbf{Decoding / seed policy} \\
\midrule
M0 & 1 & Original AI text & Deterministic \\
M1 & 1 & Qwen/\allowbreak Qwen3-4B-Instruct-2507 via vLLM & Temp. 0.9, top-p 0.95; vLLM default engine seed \\
M2 (ours) & 1 & StealthRL LoRA sampler on Qwen/\allowbreak Qwen3-4B-Instruct-2507 & Temp. 1.0, top-p 0.9; original run plus repeat seeds 4201/4202 \\
M3 & 4 & Qwen/\allowbreak Qwen3-4B-Instruct-2507 via vLLM & Temp. 1.0, top-p 0.95; vLLM default engine seed \\
M4 & 1 & \texttt{authormist/}\allowbreak\texttt{authormist-originality} via vLLM & Temp. 0.7, top-p 0.9; vLLM default engine seed \\
M5 & 1 & SilverSpeak homoglyph transform & Substitution rate 0.10; seed 42 \\
\bottomrule
\end{tabularx}
\captionof{table}{Per-method generation protocol. Table~\ref{tab:hyperparams} reports the M2 training/evaluation configuration.}
\label{tab:attack_protocol}
\end{center}

\begin{center}
\scriptsize
\setlength{\tabcolsep}{3pt}
\begin{tabularx}{\columnwidth}{l >{\raggedright\arraybackslash}X}
\toprule
\textbf{Method} & \textbf{Selection and attack-time query budget} \\
\midrule
M0 & No generation and no detector queries. \\
M1 & No reranking; 0 detector queries; 1 generation call. \\
M2 (ours) & No reranking; 0 evaluation-time detector queries; 1 policy generation call. \\
M3 & Select lowest RoBERTa score among candidates with E5 $\geq 0.90$; fallback to lowest RoBERTa score; 4 RoBERTa queries and 4 E5 scores per sample. \\
M4 & No reranking; 0 detector queries; 1 generation call. \\
M5 & Deterministic character substitution; 0 detector queries and 0 LLM calls. \\
\bottomrule
\end{tabularx}
\captionof{table}{Per-method selection and query budget. The budget counts detector calls used to construct an attack output, not the post-hoc evaluation pass in which every final output is scored once by each evaluation detector.}
\label{tab:attack_protocol_queries}
\end{center}

\FloatBarrier
\newpage
\section{Hyperparameters and Configuration}
\label{sec:hyperparams}

\begin{center}
\footnotesize
\setlength{\tabcolsep}{3pt}
\resizebox{\columnwidth}{!}{%
\begin{tabular}{ll}
\toprule
\textbf{Parameter} & \textbf{Value} \\
\midrule
\multicolumn{2}{l}{\textit{Model \& LoRA}} \\
Base model & Qwen/Qwen3-4B-Instruct-2507 \\
LoRA rank & 32 \\
LoRA alpha & 32 \\
LoRA dropout & 0.05 \\
\midrule
\multicolumn{2}{l}{\textit{Training}} \\
Algorithm & Group-Relative Policy Optimization (GRPO) \\
Learning rate & $2.8 \times 10^{-4}$ \\
Batch size & 16 \\
Group size & 8 \\
Epochs & 3 \\
Training samples & 10,000 AI-generated (custom MAGE subset) \\
KL penalty coefficient & 0.05 \\
Reference policy & Qwen3-4B-Instruct (frozen) \\
\midrule
\multicolumn{2}{l}{\textit{Reward}} \\
Detector weight ($\alpha$) & 1.0 \\
Semantic weight ($\beta$) & 0.1 \\
Detector ensemble & RoBERTa (0.6) + Fast-DetectGPT (0.4) \\
Semantic metric & E5 embedding cosine similarity \\
\midrule
\multicolumn{2}{l}{\textit{Inference}} \\
Temperature & 1.0 \\
Top-p & 0.9 \\
Max tokens & 512 \\
Prompt template & ``Paraphrase the following text while \\
& preserving its meaning: [TEXT]'' \\
\midrule
\multicolumn{2}{l}{\textit{Detectors}} \\
RoBERTa OpenAI & openai-community/roberta-large-openai-detector \\
Fast-DetectGPT & Scoring model: EleutherAI/gpt-neo-2.7B \\
Binoculars & Lightweight: gpt2-medium + gpt2-large (held-out) \\
MAGE detector & Longformer classifier (\texttt{yaful/MAGE}, held-out) \\
\midrule
\multicolumn{2}{l}{\textit{Evaluation}} \\
Test samples & 15,310 human + 14,656 AI (filtered MAGE test) \\
Token window & 100--500 tokens \\
FPR calibration & 1\% on 15,310 human samples (quantile) \\
M2 candidates per sample & 1 (see Table~\ref{tab:attack_protocol} for M0--M5) \\
\midrule
\multicolumn{2}{l}{\textit{Compute}} \\
Training framework & Tinker API (Thinking Machines) \\
Reward computation & 8$\times$ NVIDIA RTX A5000 GPUs \\
Offline generation/evaluation & 8$\times$ NVIDIA RTX A5000 GPUs \\
Execution mode & Local GPU jobs; detector scoring distributed \\
& across A5000 workers \\
Software & Python 3.12.3, PyTorch 2.10.0, Transformers 4.57.6, \\
& vLLM 0.17.1 \\
Seed & 42 \\
\bottomrule
\end{tabular}
}
\captionof{table}{Complete hyperparameters and configuration for reproducibility.}
\label{tab:hyperparams}
\end{center}

\FloatBarrier
\section{Model-Selection Ablations}
\label{app:ablations}

The main results use the full filtered MAGE test pool (15,310 human / 14,656 AI). Before running that expensive evaluation, we used smaller ablations for model selection. These ablations are diagnostic: they use smaller subsets, fewer detectors, or training-dashboard metrics, and are therefore not directly comparable to the full-test numbers in Tables~\ref{tab:app_main_binoc}--\ref{tab:app_main_part2}.

Table~\ref{tab:reward_ablation} reports the initial reward-weight sweep. All runs use RoBERTa and Fast-DetectGPT as the training detectors and are evaluated with the training-loop detector score rather than the final low-FPR detector panel. The sweep identifies $\alpha{=}1.0,\beta{=}0.1$ as the best balanced setting: it has the lowest detector probability and highest evasion rate while retaining high E5 similarity. Increasing the semantic weight to 0.3 does not improve fidelity and weakens evasion, while removing detector reward makes the objective less aligned with the attack goal.

\begin{center}
\small
\setlength{\tabcolsep}{6pt}
\resizebox{\columnwidth}{!}{%
\begin{tabular}{lcccc}
\toprule
\textbf{Reward setting} & \textbf{Detector prob.} & \textbf{Evasion rate} & \textbf{E5 Sim.} & \textbf{Total reward} \\
\midrule
$\alpha{=}1.0,\beta{=}0.0$ & 0.331 & 0.669 & 0.978 & -0.136 \\
$\alpha{=}1.0,\beta{=}0.1$ (selected) & 0.226 & 0.774 & 0.970 & 0.127 \\
$\alpha{=}1.0,\beta{=}0.3$ & 0.327 & 0.673 & 0.968 & 0.200 \\
$\alpha{=}0.0,\beta{=}1.0$ & 0.359 & 0.641 & 0.964 & 0.970 \\
\bottomrule
\end{tabular}
}
\captionof{table}{Training-time reward-weight ablation on the small MAGE development setup. Detector probability is the mean training-ensemble probability of AI generation, so lower detector probability and higher evasion rate are better for the attacker. This diagnostic sweep selected the nonzero but small semantic weight $\beta{=}0.1$.}
\label{tab:reward_ablation}
\end{center}

Table~\ref{tab:ablation_selection} reports the later checkpoint-selection evaluation on a fixed 1,000-human / 1,000-AI MAGE selection set. This pass uses the same low-FPR calibration procedure as the main evaluation, but only three detectors (RoBERTa, Fast-DetectGPT, and Binoculars); MAGE was kept for the final full-panel evaluation. The selected $\alpha{=}1.0,\beta{=}0.1$ checkpoint improves low-FPR evasion and generation quality relative to the no-semantic-reward checkpoint, and the replicate shows the same quality-preserving trend.

\begin{center}
\small
\setlength{\tabcolsep}{3pt}
\resizebox{\columnwidth}{!}{%
\begin{tabular}{lcccccc}
\toprule
\textbf{Selection run} & \textbf{Mean AUROC} & \textbf{Mean TPR@1\%FPR} & \textbf{Mean ASR} & \textbf{E5 Sim.} & \textbf{PPL} & \textbf{Judge quality} \\
\midrule
No semantic reward / low-fidelity checkpoint & 0.460 & 0.089 & 0.911 & 0.909 & 91.5 & 1.63 \\
$\alpha{=}1.0,\beta{=}0.1$ (selected) & 0.546 & 0.050 & 0.950 & 0.973 & 31.7 & 4.10 \\
$\alpha{=}1.0,\beta{=}0.1$ (replicate) & 0.554 & 0.063 & 0.937 & 0.974 & 30.5 & -- \\
\bottomrule
\end{tabular}
}
\captionof{table}{Checkpoint-selection ablations on a fixed 1,000-human / 1,000-AI MAGE selection set. Metrics are averaged across RoBERTa, Fast-DetectGPT, and Binoculars; lower AUROC and TPR are better for the attacker, while higher ASR and quality metrics are better. The full paper results are recomputed separately on the full filtered MAGE test pool.}
\label{tab:ablation_selection}
\end{center}

We also ran a candidate-budget sweep for the selected M2 policy. Increasing candidates from 1 to 2, 4, and 8 raised mean ASR from 0.997 to 1.000 on the small selection setup, while E5 similarity remained in a narrow 0.898--0.904 band. We therefore report the stricter single-shot setting in the main paper: one generation per input, with no test-time detector-guided search or reranking.

\FloatBarrier
\section{Per-Detector Results with Confidence Intervals}
\label{app:perdetector}

Tables~\ref{tab:perdetector_roberta}--\ref{tab:perdetector_mage} provide complete per-detector, per-method results with 95\% bootstrap confidence intervals (500 iterations). We split the table by detector to keep confidence intervals legible in the two-column format.

\begin{center}
\footnotesize
\setlength{\tabcolsep}{4pt}
\resizebox{\columnwidth}{!}{%
  \begin{tabular}{lccc}
\toprule
Method & AUROC [95\% CI] & TPR@1\%FPR [95\% CI] & ASR [95\% CI] \\
\midrule
M0 No Attack & 0.806 [0.801, 0.810] & 0.203 [0.196, 0.209] & 0.797 [0.791, 0.804] \\
M1 Simple Para. & 0.778 [0.773, 0.783] & 0.111 [0.106, 0.116] & 0.889 [0.884, 0.894] \\
M2 StealthRL (Ours) & 0.691 [0.685, 0.697] & 0.002 [0.002, 0.003] & 0.998 [0.997, 0.998] \\
M3 Adv.\ Para. & 0.651 [0.645, 0.657] & 0.058 [0.055, 0.062] & 0.942 [0.938, 0.945] \\
M4 AuthorMist & \textbf{0.627} [0.620, 0.633] & 0.056 [0.052, 0.059] & 0.944 [0.941, 0.948] \\
M5 Homoglyph & 0.630 [0.624, 0.636] & \textbf{0.001} [0.000, 0.001] & \textbf{0.999} [0.999, 1.000] \\
\bottomrule
\end{tabular}

}
\captionof{table}{RoBERTa results with 95\% bootstrap confidence intervals. Bold indicates the best attacker value per metric.}
\label{tab:perdetector_roberta}
\end{center}

\begin{center}
\footnotesize
\setlength{\tabcolsep}{4pt}
\resizebox{\columnwidth}{!}{%
  \begin{tabular}{lccc}
\toprule
Method & AUROC [95\% CI] & TPR@1\%FPR [95\% CI] & ASR [95\% CI] \\
\midrule
M0 No Attack & 0.661 [0.655, 0.668] & 0.388 [0.380, 0.396] & 0.612 [0.604, 0.620] \\
M1 Simple Para. & 0.588 [0.582, 0.595] & 0.148 [0.143, 0.154] & 0.852 [0.846, 0.857] \\
M2 StealthRL (Ours) & 0.089 [0.086, 0.092] & 0.002 [0.001, 0.002] & 0.998 [0.998, 0.999] \\
M3 Adv.\ Para. & 0.536 [0.530, 0.543] & 0.092 [0.087, 0.097] & 0.908 [0.903, 0.913] \\
M4 AuthorMist & 0.518 [0.511, 0.524] & 0.067 [0.064, 0.072] & 0.933 [0.928, 0.936] \\
M5 Homoglyph & \textbf{0.080} [0.078, 0.083] & \textbf{0.000} [0.000, 0.000] & \textbf{1.000} [1.000, 1.000] \\
\bottomrule
\end{tabular}

}
\captionof{table}{Fast-DetectGPT results with 95\% bootstrap confidence intervals. Bold indicates the best attacker value per metric.}
\label{tab:perdetector_fastdetectgpt}
\end{center}

\begin{center}
\footnotesize
\setlength{\tabcolsep}{4pt}
\resizebox{\columnwidth}{!}{%
  \begin{tabular}{lccc}
\toprule
Method & AUROC [95\% CI] & TPR@1\%FPR [95\% CI] & ASR [95\% CI] \\
\midrule
M0 No Attack & 0.705 [0.699, 0.712] & 0.367 [0.360, 0.375] & 0.633 [0.625, 0.640] \\
M1 Simple Para. & 0.593 [0.587, 0.599] & 0.158 [0.152, 0.163] & 0.842 [0.837, 0.848] \\
M2 StealthRL (Ours) & \textbf{0.055} [0.052, 0.058] & 0.002 [0.001, 0.003] & 0.998 [0.997, 0.999] \\
M3 Adv.\ Para. & 0.511 [0.504, 0.517] & 0.051 [0.048, 0.055] & 0.949 [0.945, 0.952] \\
M4 AuthorMist & 0.559 [0.553, 0.566] & 0.036 [0.033, 0.038] & 0.964 [0.962, 0.967] \\
M5 Homoglyph & 0.593 [0.587, 0.599] & \textbf{0.000} [0.000, 0.000] & \textbf{1.000} [1.000, 1.000] \\
\bottomrule
\end{tabular}

}
\captionof{table}{Binoculars results with 95\% bootstrap confidence intervals. Bold indicates the best attacker value per metric.}
\label{tab:perdetector_binoculars}
\end{center}

\begin{center}
\footnotesize
\setlength{\tabcolsep}{4pt}
\resizebox{\columnwidth}{!}{%
  \begin{tabular}{lccc}
\toprule
Method & AUROC [95\% CI] & TPR@1\%FPR [95\% CI] & ASR [95\% CI] \\
\midrule
M0 No Attack & 0.983 [0.981, 0.984] & 0.843 [0.837, 0.849] & 0.157 [0.151, 0.163] \\
M1 Simple Para. & 0.973 [0.972, 0.975] & 0.720 [0.713, 0.727] & 0.280 [0.273, 0.287] \\
M2 StealthRL (Ours) & 0.891 [0.888, 0.895] & \textbf{0.089} [0.084, 0.093] & \textbf{0.911} [0.907, 0.916] \\
M3 Adv.\ Para. & 0.967 [0.965, 0.969] & 0.666 [0.658, 0.674] & 0.334 [0.326, 0.342] \\
M4 AuthorMist & 0.969 [0.967, 0.971] & 0.651 [0.644, 0.658] & 0.349 [0.342, 0.356] \\
M5 Homoglyph & \textbf{0.772} [0.767, 0.777] & 0.136 [0.130, 0.141] & 0.864 [0.859, 0.870] \\
\bottomrule
\end{tabular}

}
\captionof{table}{MAGE detector results with 95\% bootstrap confidence intervals. Bold indicates the best attacker value per metric.}
\label{tab:perdetector_mage}
\end{center}

\FloatBarrier
\section{LLM Judge Prompt Templates}
\label{app:prompts}

We use separate gpt-5-nano prompts for quality and semantic similarity. Method labels are never shown to the judge. The exact quality-rating system prompt is:

\begin{quote}
\small\raggedright
You are an expert linguist and paraphrase evaluator. Your task is to assess the overall writing quality of a paraphrased text compared to the original source text.

Judge the paraphrase as real writing, not as an attack artifact. Reward unusual sentence structure only when it still reads naturally. Penalize awkward clause order, stilted transitions, filler phrases, obvious prompt leakage, parenthetical meta-comments, detector-evasion artifacts, and phrasing that feels unnatural even if the meaning is preserved.

Be strict about naturalness. Use the full 1-5 scale and avoid defaulting to 3 unless it truly fits.

Scoring criteria:
5 - Excellent: Fluent, grammatical, and natural throughout; reads like polished human writing.
4 - Good: Clear and mostly natural with only minor awkwardness.
3 - Acceptable: Understandable, but noticeably stiff, awkward, or uneven.
2 - Poor: Readably degraded; unnatural syntax or phrasing frequently distracts.
1 - Unusable: Severely broken, incoherent, or clearly unnatural.

Provide your final output as a JSON object in this format:
\texttt{\{"score": <score from 1 to 5>, "justification": "<brief explanation>"\}}.
\end{quote}

The quality-rating user prompt template is:

\begin{quote}
\small\raggedright
Evaluate the following paraphrase using the criteria above (focus on fluency and overall writing quality):

Original Text:
\texttt{"""\{original\_text\}"""}

Paraphrased Text:
\texttt{"""\{paraphrased\_text\}"""}

What score (1 to 5) would you assign to the paraphrase's quality, and why? Do not give extra credit for sounding evasive or unusual if the prose becomes less natural.
\end{quote}

The exact semantic-similarity system prompt is:

\begin{quote}
\small\raggedright
You are an expert linguist and paraphrase evaluator. Your task is to assess how well the paraphrased text preserves the meaning of the original source text.

Be generous when meaning is substantially preserved. Use the full 1-5 scale and avoid defaulting to 3 unless it truly fits. If the paraphrase keeps the core meaning with only minor changes, prefer 4 or 5. Use 2 or 1 only for clear meaning loss or topic drift.

Scoring criteria:
5 - Approximately equivalent: Meaning preserved; only wording/structure changes.
4 - Nearly equivalent: Meaning mostly preserved; small factual or emphasis shifts.
3 - Somewhat equivalent: Partial meaning preserved; important details differ.
2 - Topically related: Same topic but most meaning lost.
1 - Not related: Different topic or meaning.

Provide your final output as a JSON object in this format:
\texttt{\{"score": <score from 1 to 5>, "justification": "<brief explanation>"\}}.
\end{quote}

The semantic-similarity user prompt template is:

\begin{quote}
\small\raggedright
Evaluate the following paraphrase using the criteria above:

Original Text:
\texttt{"""\{original\_text\}"""}

Paraphrased Text:
\texttt{"""\{paraphrased\_text\}"""}

What score (1 to 5) would you assign for semantic similarity, and why?
\end{quote}

\section{Dataset Statistics}
\label{app:dataset}

\begin{center}
\footnotesize
\setlength{\tabcolsep}{3pt}
\resizebox{\columnwidth}{!}{%
\begin{tabular}{llcc}
\toprule
\textbf{Stage} & \textbf{Purpose} & \textbf{Human Samples} & \textbf{AI Samples} \\
\midrule
Train           & RL fine-tuning (GRPO)       & 0 & 10,000 \\
Test            & Detection evaluation        & 15,310  & 14,656  \\
Likert eval     & Quality scoring (per method) & 0     & 500    \\
\midrule
Token range     & \multicolumn{3}{c}{100--500 tokens} \\
Source          & \multicolumn{3}{c}{MAGE benchmark~\cite{li2024magemachinegeneratedtextdetection}} \\
Domains         & \multicolumn{3}{c}{Multiple (essays, Q\&A, creative writing)} \\
\bottomrule
\end{tabular}
}
\captionof{table}{Dataset statistics for training and evaluation. The Likert evaluation uses 500 AI samples per attack method (M1--M5) with matched sample IDs.}
\label{tab:dataset}
\end{center}

\FloatBarrier
\section{Supplementary Figures}
\label{app:figures}

\begin{figure*}[t]
  \centering
  \includegraphics[width=\textwidth]{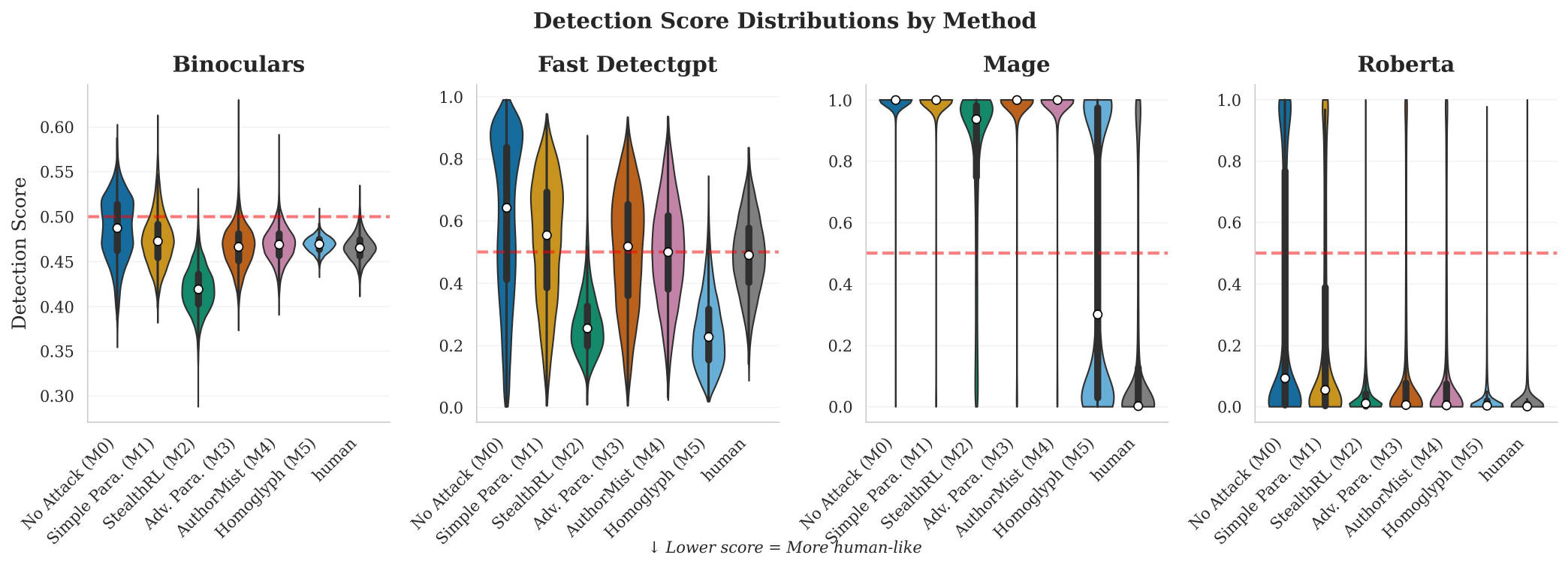}
  \caption{Detector score distributions for AI samples across methods (one panel per detector). StealthRL (M2) and Homoglyph (M5) push scores below the detection threshold, explaining their near-zero TPR@1\%FPR.}
  \label{fig:app_score_dist}
\end{figure*}

\begin{figure*}[t]
  \centering
  \includegraphics[width=0.75\textwidth]{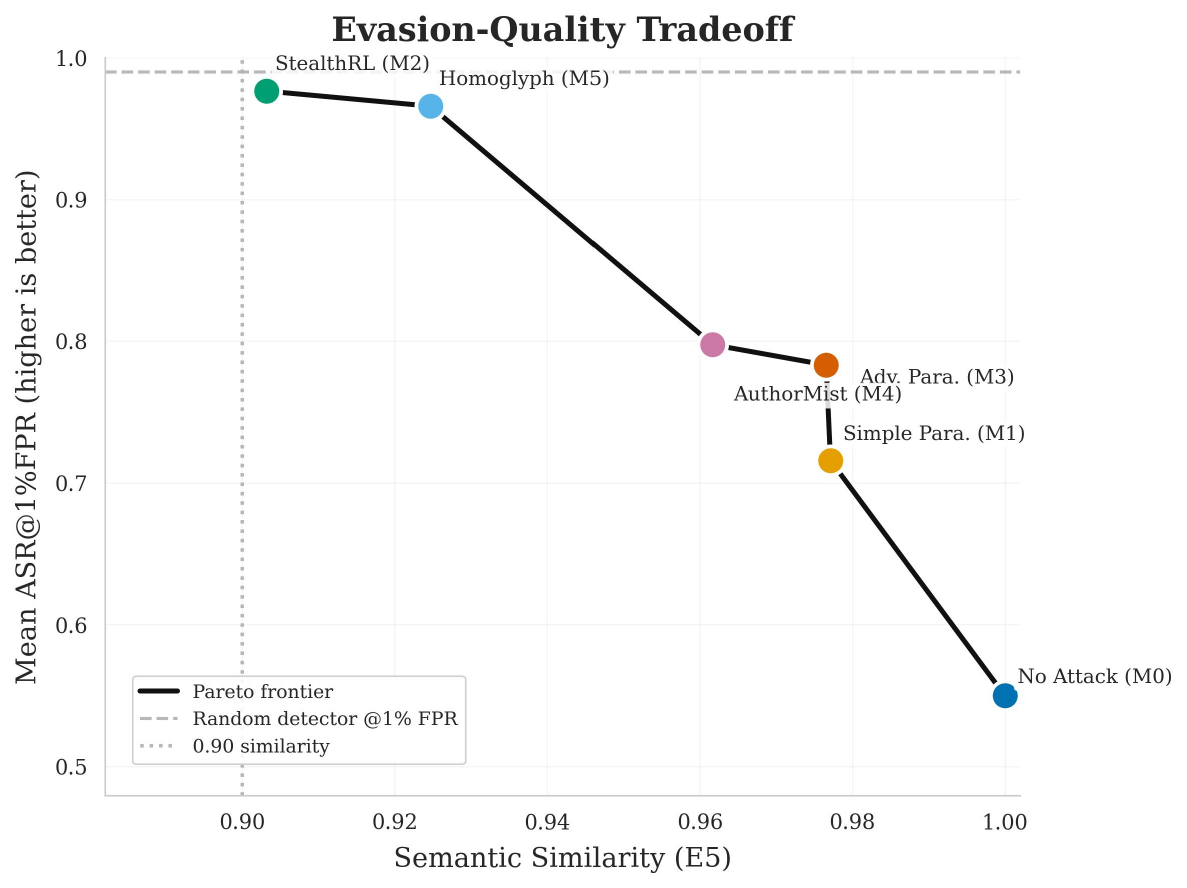}
  \caption{Evasion--quality tradeoff at the 1\% FPR operating point. Each point represents a method, plotted by E5 semantic similarity (x-axis) against mean ASR@1\%FPR (y-axis). Top-right is ideal (high similarity, high attack success). The dashed lines mark the 0.90 similarity floor and the random-detector ASR baseline at 1\% FPR, and the solid line shows the Pareto frontier.}
  \label{fig:app_tradeoff}
\end{figure*}

\begin{figure*}[t]
  \centering
  \includegraphics[width=0.62\textwidth]{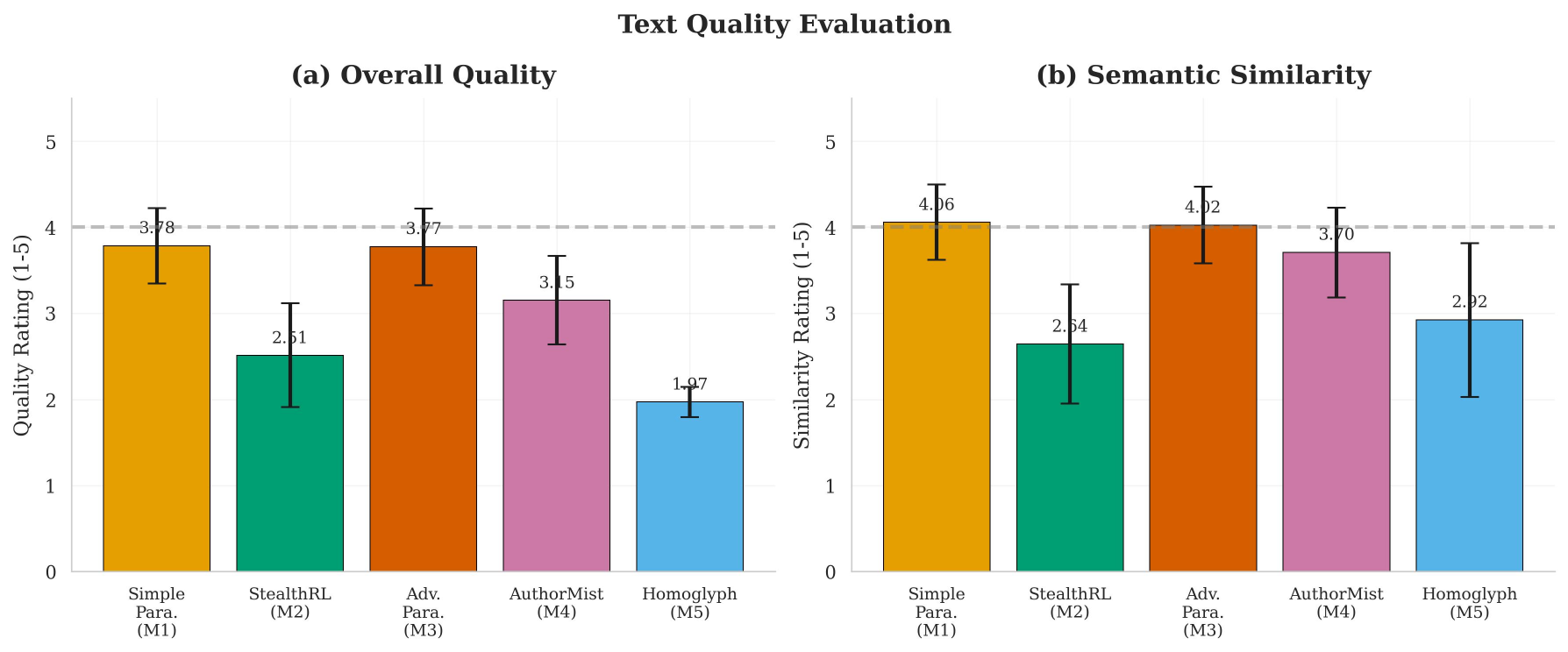}
  \caption{LLM-based quality evaluation on 500 matched samples. Panel (a) shows linguistic quality and panel (b) semantic similarity; higher is better. The dashed line marks the neutral midpoint (3.0).}
  \label{fig:app_quality}
\end{figure*}

\end{document}